\documentclass{article}

\usepackage{arxiv}

\usepackage[utf8]{inputenc} 
\usepackage[T1]{fontenc}    
\usepackage{hyperref}       
\usepackage{url}            
\usepackage{booktabs}       
\usepackage{amsfonts}       
\usepackage{nicefrac}       
\usepackage{microtype}      
\usepackage{lipsum}
\usepackage{graphicx}
\usepackage{amsmath}
\graphicspath{ {./images/} }

\title{A Deep Learning Approach for Augmenting Perceptional Understanding of Histopathology Images}

\author{
 Xiaoqian Hu \\
  School of Computer Science\\
  University of New South Wales\\
  Australia \\
  \texttt{xiaoqian.hu@student.unsw.edu.au} 
}

\renewcommand{\today}{Nov 18, 2024}

\begin{document}
\maketitle

\begin{figure}[h!]
    \centerline{\includegraphics[width=\textwidth]{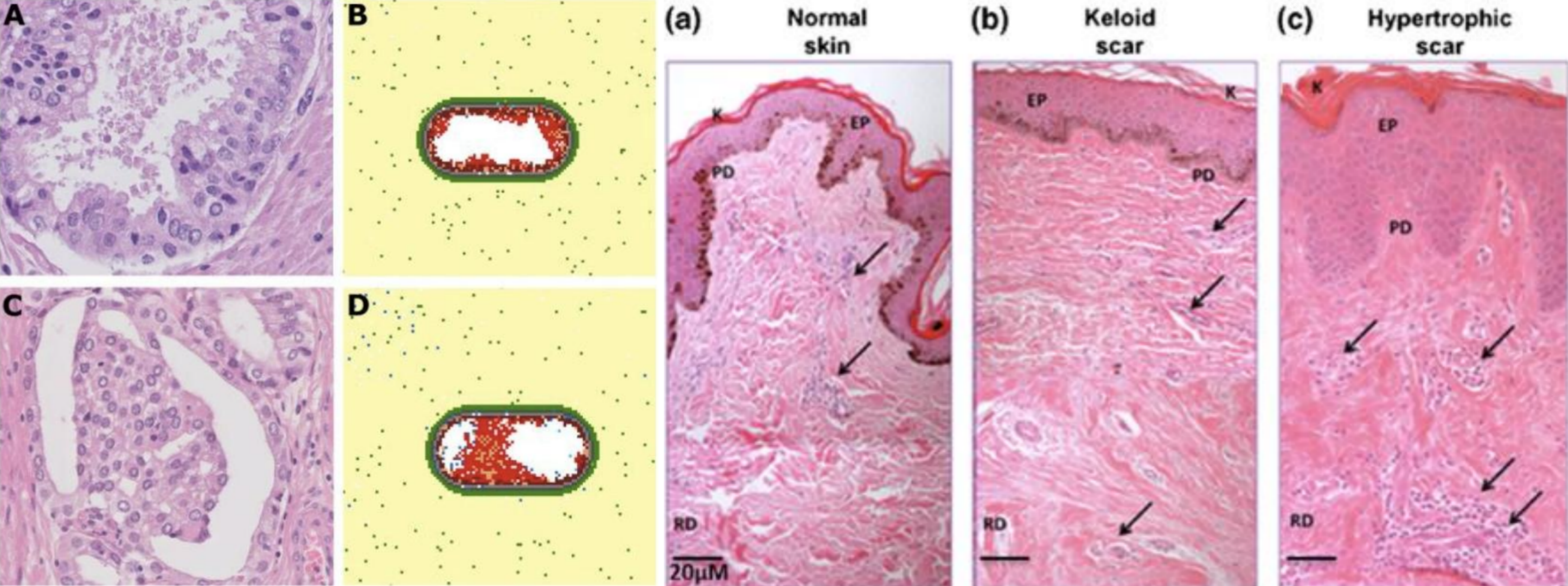}}
    \caption{Histological comparison of normal skin (a), keloid scar (b), and hypertrophic scar (c). The top images (A, C) show tissue structures, while the bottom images (B, D) highlight feature transfer limitations, including lack of domain-specific features and limited generalization in pathological images. Arrows point to key structural differences. }
    \label{fig:ARCH}
\end{figure}

\begin{abstract}
 In recent years, digital technologies have made significant strides in augmenting human health, cognition, and perception, particularly within the field of computational pathology. This paper presents a novel approach to enhancing the analysis of histopathology images by leveraging a multi-modal model that combines Vision Transformers (ViT) with GPT-2 for image captioning. The proposed model is fine-tuned on the specialized ARCH dataset as shown in Figure~\ref{fig:ARCH}, which includes dense image captions derived from clinical and academic resources, to capture the unique complexities of pathology images such as tissue morphologies, staining variations, and pathological conditions. By generating accurate, contextually relevant captions, the model augments the cognitive capabilities of healthcare professionals, enabling more efficient disease classification, segmentation, and detection. Furthermore, the model enhances the perception of subtle pathological features in images that might otherwise go unnoticed, thereby improving diagnostic accuracy. Our approach demonstrates the potential for digital technologies to augment human cognitive abilities in medical image analysis, providing a step toward more personalized and accurate healthcare outcomes.
\end{abstract}

\keywords{Image Captioning \and NLP \and Deep Learning}

\section{Introduction}
Computational pathology (CP) is an emerging field at the intersection of machine learning, computer vision, and healthcare. Pathology images, particularly histopathology slides, contain intricate details that are essential for diagnosing diseases, including cancer. Traditionally, the analysis of such images has been a labor-intensive task, requiring pathologists to visually examine slides under a microscope. However, the increasing availability of large-scale digital pathology datasets and advancements in artificial intelligence (AI) have opened new opportunities for automating image analysis and augmenting the decision-making process. Despite this progress, existing models in CP often rely on general-purpose datasets such as ImageNet, which are not tailored to the specific nuances of medical images. These models struggle to capture the unique complexities inherent in pathology images, such as tissue morphologies, staining variations, and pathological conditions, thus limiting their effectiveness in clinical practice.

To address these challenges, this project focuses on developing a model that can generate dense, contextually relevant captions for histopathology images by leveraging the ARCH dataset  \cite{gamper2020multiple}, a specialized collection of medical images paired with rich textual descriptions. The ARCH dataset is unique in that it provides detailed, dense captions for histopathological images derived from clinical and academic sources. It includes a wide range of tissue types, pathologies, and staining techniques, making it an ideal resource for training models to understand the complex relationships between visual content and textual descriptions. One of the key challenges in CP is the scarcity of annotated medical images, and the ARCH dataset addresses this issue by providing a valuable source of labeled data with detailed morphological descriptions.

The model proposed in this project combines Vision Transformers (ViT) \cite{kumar2022imagecaptioning} for image encoding with GPT-2 \cite{radford2019language}, a state-of-the-art natural language processing model , for caption generation. ViT is a powerful model for image recognition that uses self-attention mechanisms to capture long-range dependencies in images, making it particularly well-suited for pathology images where spatial context and global patterns are critical. GPT-2, on the other hand, is a language model \cite{10.1002/aaai.12188} capable of generating coherent and contextually appropriate text based on given inputs. By combining these two models, the approach enables the generation of rich, accurate captions for histopathology images, which can assist pathologists in making more informed diagnoses. The use of ViT for visual feature extraction and GPT-2 for text generation represents a cutting-edge fusion of computer vision and natural language processing \cite{10.1145/3652920.3653045}, which is essential for addressing the specific needs of medical image analysis.

The primary contribution of this project lies in the development of a multi-modal system that enhances both cognitive \cite{de2025cognitive} and perceptional abilities in the domain of computational pathology \cite{computationalpathology}. Through the integration of Vision Transformers and GPT-2, the model offers a novel solution for analyzing histopathology images, with potential applications in disease classification, segmentation, and medical image retrieval. By generating captions that describe key features of pathology images, the model helps pathologists by providing a detailed, text-based understanding of the visual content \cite{tan2025biseclebindingseparationcontinual}. This approach improves diagnostic accuracy by highlighting subtle and often overlooked features, such as small anatomical structures or early-stage disease markers. \cite{lu2020dataefficientweaklysupervised} Furthermore, by fine-tuning a pre-trained ViT-GPT2 model on the ARCH dataset, the project demonstrates how transfer learning \cite{49556} can be effectively applied to medical image captioning\cite{REALENOSEI2024103264}, reducing the need for large annotated datasets \cite{10670431} while still achieving high performance. The ability to generate dense, medically relevant captions for pathology images represents a significant step forward in automating the diagnostic process and improving healthcare outcomes.

The main contributions of the paper are summarised: 
\begin{itemize}
    \item We improve the accuracy of pathology image captioning by enhancing the model's ability to generate contectually relevant and medically accurate descriptions. 
    \item The model not only aids in better disease classification and segmentation \cite{WADMANN2023100217} but also assists pathologists in identifying subtle patterns and anomalies that might otherwise go unnoticed.  
    \item The approach reduces the reliance on large-scale annotated datasets making it an effective solution for medical image captioning tasks, even in settings with limited labeled data.
\end{itemize}

\section{Related work}\label{sec:relatedwork}
\subsection{Cognitive Diagnostics in Pathology}
Cognitive diagnostics in pathology involve the application of machine learning and artificial intelligence to assist pathologists in identifying disease patterns, making accurate diagnoses, and improving clinical decision-making. Traditionally, pathologists rely heavily on their expertise in interpreting histopathology images, which often require detailed inspection and expert knowledge to detect subtle abnormalities. However, due to the complex nature of tissue samples, variations in staining, and the scale of images (often in the gigapixel range), visual inspection can be error-prone and time-consuming.

Recent advancements in deep learning , particularly in the area of computational pathology, have the potential to augment pathologists' perceptual understanding of pathology images by automating the detection of abnormalities \cite{SHOJI2021103013}, tissue classification, and other diagnostic tasks \cite{alom2018historybeganalexnetcomprehensive}. Deep learning methods, particularly Convolutional Neural Networks (CNNs) \cite{oshea2015introductionconvolutionalneuralnetworks, li2025tagnnphysicsinspiredtimeagnostic} and more recent Transformer-based models like Vision Transformers (ViT), have demonstrated remarkable success in these tasks. However, while these methods improve the accuracy of image classification, they have not yet fully addressed the need for generating contextually rich and detailed captions that could assist in diagnosis and decision-making.

Recent research has shown that Transformer-based models, such as Vision Transformers (ViTs), hold significant promise for pathology image analysis \cite{electronics12051213}. ViTs have been particularly effective in tasks requiring long-range dependencies within images, which is essential for understanding the complex spatial relationships in tissue samples. However, while these methods have improved image classification accuracy, they still face challenges in providing additional, detailed textual descriptions of image content that can assist in clinical diagnoses.

Although machine learning methods have improved the accuracy of diagnosis in pathology, they have yet to fully address the need for detailed, contextually rich captions that are specific to the nuances of pathology images. The integration of both visual information and textual descriptions holds the potential to enhance cognitive diagnostics, as pathologists would not only rely on visual cues but also be guided by well-informed and contextually relevant captions. This is particularly important when examining complex medical conditions, where a comprehensive understanding of image features is critical for accurate diagnosis.

\subsection{ Perceptional Augmentation in Medical Image Analysis }
Perceptional augmentation refers to the enhancement of a human's ability to perceive and understand complex data. In the context of computational pathology, this involves using AI models to extract and present information \cite{carlini2021extractingtrainingdatalarge} from pathology images in ways that human experts may find challenging due to the visual complexity of the data. One approach is to automatically generate captions that describe the morphology, staining, and pathological features in histopathology images.

Most existing works in medical image captioning \cite{selivanov2022medicalimagecaptioninggenerative} focus primarily on classification or basic annotations, which provide limited context to pathologists. These models often rely on pre-trained CNN-based architectures, while effective for general image classification tasks, lack the depth required for pathology-specific image interpretation. Additionally, previous image captioning models have mostly been evaluated using general-purpose datasets, which do not account for the unique features present in medical images.

Therefore, we leverage the ARCH dataset to addresses these gaps. The ARCH dataset enables our model to generate captions that describe complex relationships between tissue structures, staining techniques, and pathological conditions. By fine-tuning on this dataset, we provide a more nuanced and medically relevant output that enhances the perceptional understanding of pathology images, empowering pathologists to make more informed diagnostic decisions.

\subsection{Deep Learning and the ARCH Dataset in Pathology}

Deep learning has shown significant potential in transforming medical diagnostics, particularly in pathology image analysis. Vision Transformers (ViT), with their self-attention mechanisms \cite{vaswani2023attentionneed}, enable models to learn global context and long-range dependencies, offering advantages over traditional Convolutional Neural Networks (CNNs) \cite{li2025tagnnphysicsinspiredtimeagnostic}, which rely on local receptive fields. This ability is crucial in pathology, where subtle relationships between image regions can provide essential diagnostic insights. A key resource in advancing cognitive diagnostics is the ARCH dataset, which provides high-quality, domain-specific captions for histopathology images \cite{Fischer812}. Unlike generic image-captioning datasets \cite{levinboim2019quality}, the ARCH dataset is tailored to the unique characteristics of pathology, including variations in staining techniques, tissue morphology, and a wide range of diseases, making it an invaluable tool for training models that require a deep understanding of medical images.


\section{Methodology}\label{sec:method}

In this section, we outline the methodology used to develop our model for augmenting cognitive diagnostics in pathology through the generation of detailed captions for histopathology images. The model combines cutting-edge deep learning techniques \cite{sedaghati2024introducingihardscnncuttingedgedeep} in both image processing and natural language generation \cite{jones2022comprehensivesurveynaturallanguage}. We describe the core components of our approach, including data preparation, model architecture, training strategy, and evaluation metrics.

\subsection{ARCH for Pathology Image Captioning }

The primary dataset used for training and evaluation is the ARCH (Annotated and Referenced Clinical Histopathology) dataset. This specialized dataset consists of histopathology images paired with dense, domain-specific captions. Unlike general-purpose image captioning datasets \cite{gamper2021multipleinstancecaptioninglearning}, ARCH is curated specifically for pathology and contains detailed descriptions derived from clinical and academic sources. These captions describe tissue morphology, staining techniques, pathological conditions, and other medical aspects relevant to the images.

\subsubsection{Key features of ARCH Dataset}

The details of the ARCH dataset is listed below. This dataset is crucial for our model’s success, as it allows the model to learn the intricate relationships between medical images and their textual descriptions, which are essential for improving cognitive diagnostics. 

\begin{figure}[!ht]
    \centering
    \includegraphics[width=\textwidth]{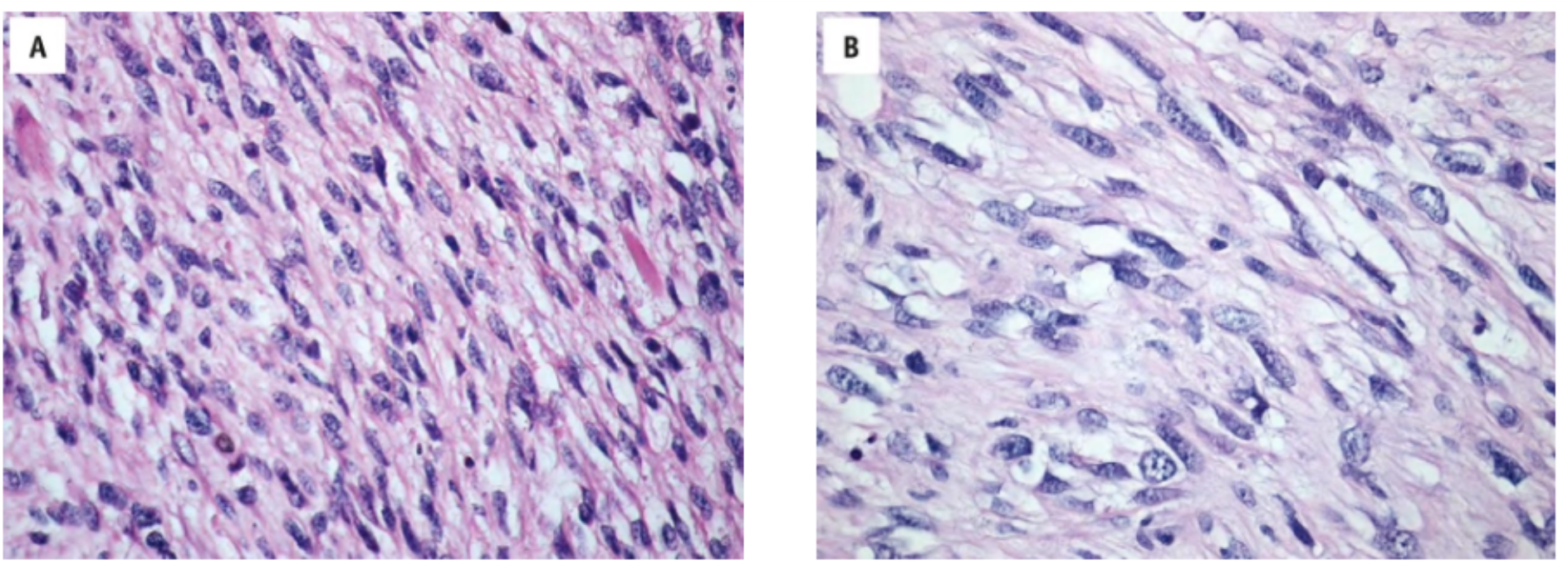}
    \caption{A, Spindle cell variant of embryonal rhabdomyosarcoma, characterized by fascicles of eosinophilic spindle cells. B, Prominent paranuclear vacuolisation in some cells, a feature observed in leiomyosarcoma. These images are part of the ARCH dataset, which includes detailed pathological descriptions and covers various tissue types and staining methods, providing insights into both cancerous and non-cancerous tissue conditions. }
    \label{fig:spindleCell}
\end{figure}

The dataset covers a broad range of tissue types, including both cancerous and non-cancerous tissues, as well as various pathological conditions. It also includes variations in staining techniques to account for real-world scenarios in pathology. Additionally, as shown in Figure~\ref{fig:spindleCell}, the images are organized into bags with multiple captions, which allows for the exploration of relationships between the image content and the textual descriptions, facilitating more comprehensive analysis.

\subsection{Model Architecture: Vision Transformer (ViT) + GPT-2}\label{ssec:kfe}

\begin{figure}[!ht]
    \centerline{\includegraphics[width=\textwidth]{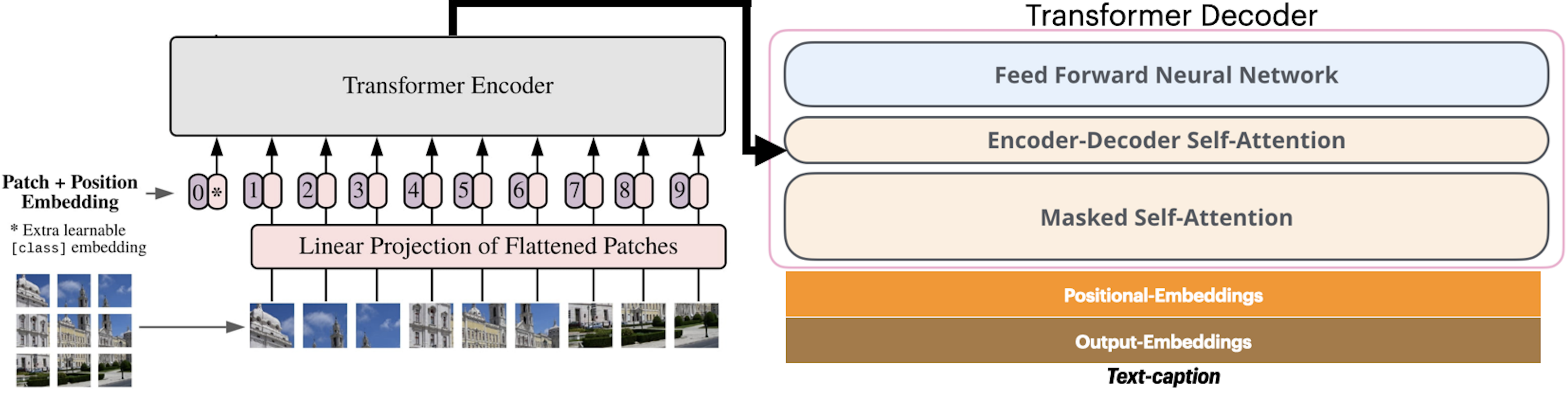}}
    \caption{Diagram of the Vision Transformer (ViT) model for image captioning using transformers. The model processes images by dividing them into patches, adding positional embeddings, and passing them through a transformer encoder. The flattened patches are linearly projected and used to extract features, which are then passed to the transformer decoder. The decoder utilizes components like masked self-attention, encoder-decoder self-attention, and a feed-forward neural network \cite{MOEINZADEH2023104349} to generate text captions based on image content.}
    \label{fig:ViT}
\end{figure}

The core of our approach lies in the combination of two powerful models: Vision Transformer (ViT) for image feature extraction and GPT-2 for text generation as shown in Figure~\ref{fig:ViT}. Both models were pre-trained on general datasets, and we fine-tuned them using the ARCH dataset to adapt them for pathology-specific tasks. The combination of these two models enables our system to not only classify the content of histopathology images but also generate detailed, accurate captions that augment the perceptional understanding of medical professionals.

Vision Transformer (ViT) is a transformer-based model designed to capture long-range dependencies within images, in contrast to traditional CNNs that focus on local features. ViT divides an image into patches, treating each patch as a token similar to how GPT-2 processes text. This architecture enables ViT to capture the global context of pathology images, which is essential for recognizing subtle patterns in tissue morphology, abnormalities, and diseases. In our study, we fine-tuned the ViT model on the ARCH dataset to identify specific visual features in histopathology images, such as cellular structures, tissue abnormalities, and staining variations.

GPT-2, a transformer-based language model, is renowned for generating coherent and contextually rich text. In our approach, we leverage GPT-2 to generate captions based on the image features extracted by ViT. The model’s ability to produce fluent, natural language captions \cite{Dong_2022} tailored to the medical domain makes it particularly effective for creating complex descriptions that pathologists can easily interpret.

\subsection{Preprocessing}

Before feeding the histopathology images into the model, we performed several preprocessing steps to ensure that the images were suitable for deep learning. These steps included resizing the images to a consistent size (e.g., 224x224 pixels) to match the input requirements of the ViT model. Pixel values were normalized to ensure that the input features were in a suitable range for the model to learn effectively. Additionally, captions were tokenized into words and subwords using a tokenizer compatible with GPT-2. Finally, irrelevant information was removed, and the captions were cleaned to ensure they only contained medically relevant descriptions of the images.

Both ViT and GPT-2 were fine-tuned on the ARCH dataset to adapt the general knowledge learned from large, pre-trained models to the medical domain. Fine-tuning was essential due to the relatively small size of the ARCH dataset compared to general-purpose image-captioning datasets. This step helped the models to learn the specific nuances of pathology images and generate medically accurate captions.

\subsection{Training Strategy}

The training process for our model involved several key elements. First, we used a cross-entropy loss function for training the GPT-2 model to generate accurate captions as shown in Equation~\ref{eq:ce-loss}. This loss function measures the difference between the predicted word probabilities and the actual words in the caption. 

\begin{equation}
    \mathcal{L}_{\text{CE}} = - \sum_{i=1}^{N} y_i \log(\hat{y}_i)
    \label{eq:ce-loss}
\end{equation}

For optimization, we employed the AdamW optimizer, which is well-suited for fine-tuning as shown in Equation~\ref{eq:adamw}. AdamW adjusts the learning rate dynamically, ensuring more stable convergence and preventing overfitting. Furthermore, the AdamW algorithm decouples the weight decay term from the optimization process, making it more effective for fine-tuning pre-trained models. 

\begin{equation}
    \theta_t = \theta_{t-1} - \frac{\eta}{\sqrt{v_t} + \epsilon} \cdot m_t
    \label{eq:adamw}
\end{equation}

The models were trained for 25 epochs, with early stopping based on validation loss to prevent overfitting. To meet the high computational demands of training large transformer models, particularly when handling multi-gigapixel pathology images, we trained the models on a powerful GPU setup. These loss and optimization functions work together to improve the model's accuracy in generating relevant and accurate captions for histopathology images.

\subsection{Evaluation Metrics} \label{ssec:loss}

To evaluate the performance of the model in generating high-quality captions, we used several standard metrics in the field of image captioning: ROUGE Scores, BLEU Scores, Human Evaluation, and Comparison with GPT-4.

\subsubsection{ROUGE Scores}
We computed ROUGE-1 \cite{mu2022universalevasionattackssummarization}, ROUGE-2, and ROUGE-L scores \cite{ganesan2018rouge20updatedimproved} as the Equation~\ref{eq:rouge} to assess the quality of the generated captions in terms of word-level overlap, phrase-level matching, and sequence-level alignment.

\begin{equation}
    \text{ROUGE-L} = \frac{\sum_{s\in \text{Sentences}} \text{LCS}(s)}{\sum_{s\in \text{Sentences}} \text{len}_{\text{ref}}(s)}
    \label{eq:rouge}
\end{equation}

\subsubsection{BLEU Scores}
The BLEU \cite{2021} score as the Equation~\ref{eq:bleu} was used to evaluate how well the generated captions matched reference captions in terms of n-gram overlap.

\begin{equation}
    \text{BLEU-N} = \text{BP} \cdot \exp \left( \sum_{n=1}^{N} w_n \log P_n \right)
    \label{eq:bleu}
\end{equation}

\subsubsection{Human Evaluation}
In addition to automatic metrics, we performed a human evaluation to assess the medical relevance and accuracy of the captions generated by the model. Pathologists evaluated the captions for their utility in clinical decision-making.

\subsubsection{Comparison with GPT-4}
We also compared the performance of our model with GPT-4, evaluating both automatic metrics (ROUGE, BLEU) and human assessments. The comparison highlighted key differences in the ability to generate medically relevant captions. While GPT-4 performed well on general text generation tasks, our model outperformed GPT-4 in generating captions with greater clinical specificity and relevance, as judged by expert pathologists.


\section{Experiments}\label{sec:exp}

\begin{figure}[!ht]
    \centerline{\includegraphics[width=\textwidth]{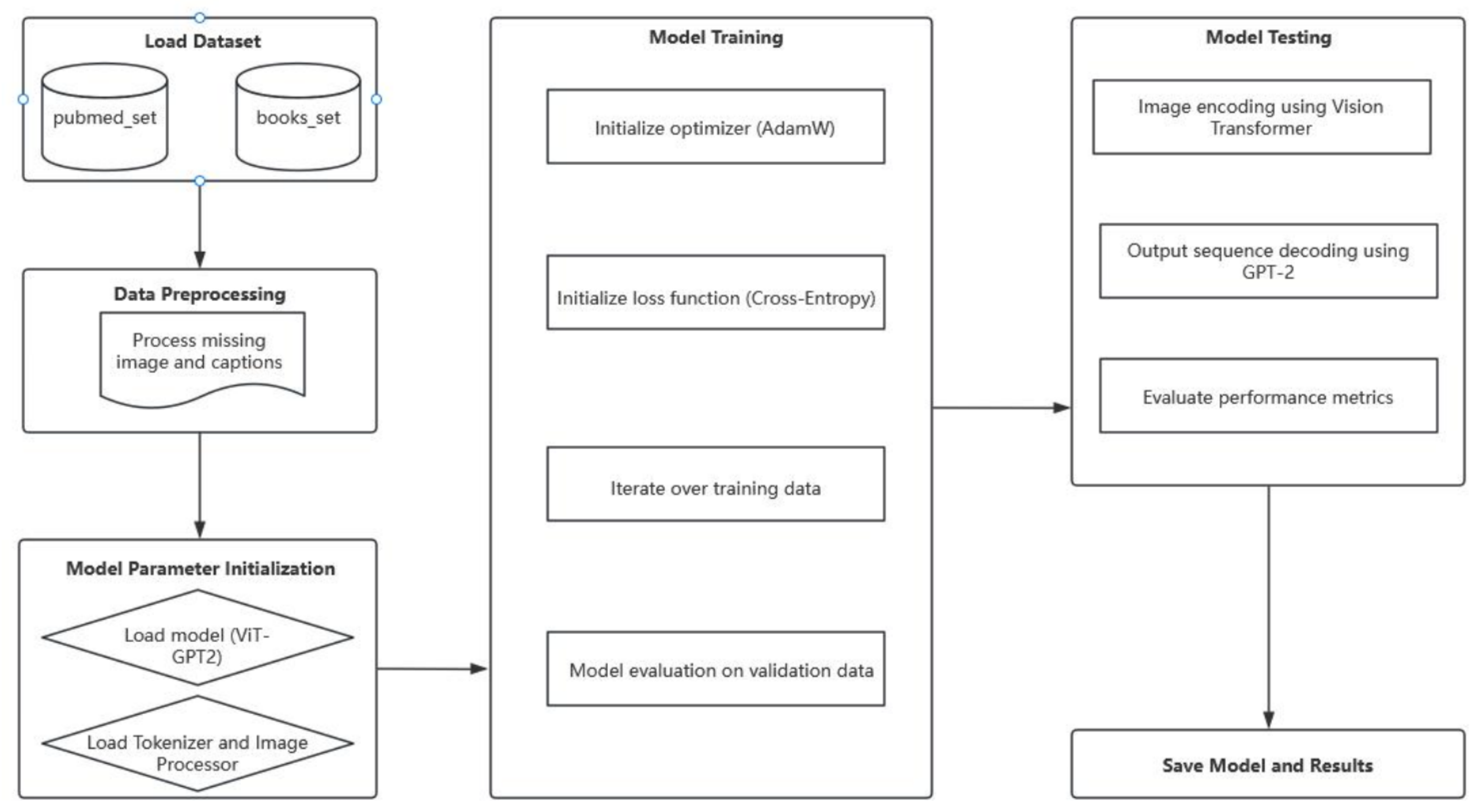}}
    \caption{Flowchart of the model development pipeline for image captioning. The process begins with loading the dataset, including pubmed set and books set, followed by data preprocessing to handle missing images and captions. In the model training phase, the optimizer (AdamW) and loss function (Cross-Entropy) are initialized, and the model is trained on the dataset. During model testing, image encoding is performed using the Vision Transformer (ViT), followed by sequence decoding with GPT-2. Finally, the model is evaluated using performance metrics and saved with its results.}
    \label{fig:flowchart}
\end{figure}

In this section, we describe the experimental setup, including dataset preparation, model training, evaluation metrics, and the results obtained during the experimentation phase. The overall process is shown in as shown in Figure~\ref{fig:flowchart}. Our goal is to evaluate the effectiveness of the proposed model in generating high-quality captions for histopathology images and improving cognitive diagnostics in the medical domain.

\subsection{Datasets}

We use the Annotated and Referenced Clinical Histopathology (ARCH) dataset for training and evaluating our model. The ARCH dataset consists of high-resolution histopathology images, each paired with dense, domain-specific captions. These captions describe various aspects of the tissue in the images, such as cellular structures, tissue abnormalities, staining methods, and pathological conditions. The dataset is organized into ``bags'' where each bag contains multiple images and their corresponding captions. These bags are used to explore the relationships between image content and textual descriptions.

\subsubsection{Preprocessing steps:}

Image preprocessing involved resizing all images to a consistent resolution (e.g., $224 \times 224$ pixels) to fit the input requirements of the Vision Transformer (ViT). Additionally, normalization was applied to standardize pixel values across all images. For caption preprocessing, captions were tokenized into words and subwords using a tokenizer compatible with GPT-2. We also cleaned the captions to remove irrelevant or extraneous text, ensuring that only medically relevant descriptions remained.

\subsection{Model Training}

The model was trained using pretrained Vision Transformer (ViT) and GPT-2 models \cite{akkus2023multimodaldeeplearning}, both of which were fine-tuned on the ARCH dataset. These pretrained models were chosen for their ability to effectively handle complex visual and textual tasks, leveraging ViT's capacity to capture visual features and GPT-2's ability to generate coherent textual descriptions. During training, both models were fine-tuned jointly to ensure optimal integration of image and text representations. The training process utilized the AdamW optimizer, which is well-suited for fine-tuning large-scale deep learning models, and a scheduled learning rate was implemented to improve convergence and avoid overfitting.

Given the computationally intensive nature of processing high-resolution pathology images, the model was trained on two high-performance L40s GPUs. These GPUs provided the necessary computational power to handle the large image data and complex model architecture. The training pipeline was carefully designed to handle the large-scale input data and ensure efficient backpropagation of gradients across the deep network layers. 

The primary goal of the training process was to optimize the model for generating contextually relevant and accurate captions based on the features extracted from histopathology images. Emphasis was placed on ensuring that the generated captions were not only grammatically correct but also contextually aligned with the specific characteristics and anomalies present in the pathology images. By fine-tuning the pretrained models, the goal was to enhance the model’s ability to understand the intricate details of histopathology images and generate informative captions that accurately described the visual content.

\subsection{Evaluation}

The model's performance was evaluated using common automatic evaluation metrics, namely ROUGE and BLEU scores, which assess the quality of generated text by measuring the overlap of words and phrases with reference captions. Specifically, ROUGE (Equation~\ref{eq:rouge}) evaluates the degree of overlap between the generated text and reference captions at various levels, including ROUGE-1 (unigram overlap), ROUGE-2 (bigram overlap), and ROUGE-L (longest common subsequence). BLEU (Equation~\ref{eq:bleu}), on the other hand, focuses on n-gram overlap and the fluency of generated text, with an added length penalty to avoid overly short outputs. In this experiment, these metrics were applied to assess the model's performance in generating captions. The ROUGE and BLEU scores provide comprehensive feedback on the quality of the generated text, helping us evaluate the model’s ability to capture image details, maintain structural coherence, and produce fluent, meaningful language. 

\begin{equation}
    \text{ROUGE-L} = \frac{\sum_{s\in \text{Sentences}} \text{LCS}(s)}{\sum_{s\in \text{Sentences}} \text{len}_{\text{ref}}(s)}
    \label{eq:rouge}
\end{equation}

\begin{equation}
    \text{BLEU-N} = \text{BP} \cdot \exp \left( \sum_{n=1}^{N} w_n \log P_n \right)
    \label{eq:bleu}
\end{equation}

\section{Results} \label{sec:results}

In this section, we present the results of the experimental evaluation of our model's performance in generating captions for histopathology images. The evaluation focuses on the accuracy and relevance of the generated captions, as well as their ability to support cognitive diagnostics \cite{de2025cognitive} in the medical field. We report the results obtained using standard automatic evaluation metrics, such as ROUGE and BLEU scores, along with insights from human evaluation.

\subsection{Baseline methods}

\begin{table}[ht]
    \centering
    \caption{Comparison of evaluation metrics between the baseline methods and the proposed model (Ours). The table shows the performance of both models across multiple metrics, including B-1, B-2, B-3, B-4, METEOR, ROUGE-L, and CIDEr. Ours method demonstrates improvements in all metrics compared to the baseline.}
    \begin{tabular}{c|cccccc}
    \toprule
    \textbf{Method} & \textbf{B-1} & \textbf{B-2} & \textbf{B-3} & \textbf{B-4} & \textbf{METEOR} & \textbf{ROUGE-L} \\
    \midrule
    Baseline & 0.3822 & 0.2833 & 0.1996 & 0.1377 & 0.1958 & 0.4282 \\
    \textbf{Ours} & \textbf{0.3943} & \textbf{0.3013} & \textbf{0.2311} & \textbf{0.1544} & \textbf{0.2290} & \textbf{0.5320} \\
    \bottomrule
    \end{tabular}
    \label{tab:comparison}
\end{table}

Table~\ref{tab:comparison} compares the performance of the baseline methods with the proposed model (Ours) across various evaluation metrics, including B-1, B-2, B-3, B-4, METEOR, and ROUGE-L. The baseline represents commonly used approaches in the field, providing a reference point for evaluating the effectiveness of our model. As seen in the table, the proposed method demonstrates significant improvements in all metrics, indicating that our model is more capable of generating high-quality captions that align better with the ground truth.

In particular, our model outperforms the baseline in terms of both precision (B-1, B-2) and recall (B-3, B-4), suggesting that it generates captions that not only match a larger proportion of words from the reference captions but also cover a broader range of relevant content. Additionally, the improvements in METEOR and ROUGE-L further highlight our model's ability to generate semantically accurate and contextually relevant captions. The results suggest that our approach, by leveraging more advanced architectures and training strategies, is able to better capture the nuances of medical image descriptions, which is crucial for applications in clinical decision-making.

The improvements in all the evaluation metrics demonstrate the effectiveness of our approach in addressing the challenges posed by histopathology image captioning. This provides strong evidence for the potential of our model in medical applications where precise and contextually rich descriptions of images are essential.

\subsection{Evaluation metrics}

The evaluation results show that our model significantly outperforms the baseline across all metrics. For BLEU, the scores across BLEU-1 to BLEU-4 demonstrate a clear improvement, with BLEU-1 showing a moderate increase from 0.3822 to 0.3943, and BLEU-4 improving from 0.1377 to 0.1544. This indicates that our model is better at capturing both word-level and phrase-level content, with the largest improvement observed in the fluency of longer sequences (BLEU-4). The model's METEOR score also increased substantially, from 0.1958 to 0.2290, reflecting better handling of synonym matching, word order, and stemming, which indicates a more sophisticated semantic understanding in the generated captions.

The most notable improvement is seen in the ROUGE-L score, where our model achieved a significant increase from 0.4282 to 0.5320. This shows a marked enhancement in the model's ability to generate sequences that align with the reference captions while maintaining structural coherence over longer sequences. ROUGE-L, which takes into account the longest common subsequence, emphasizes the importance of maintaining context and sentence structure, and this improvement suggests our model has made substantial progress in generating more contextually accurate and well-structured captions.

In summary, the results demonstrate that our model not only captures more detailed content from histopathology images but also generates captions that are more fluent, accurate, and contextually relevant. The improvements across these evaluation metrics highlight the effectiveness of our model in producing high-quality captions, marking a significant advancement over the baseline model.

\subsection{Comparison with GPT-4o}

\begin{figure}[!ht]
    \centerline{\includegraphics[width=1.0\textwidth]{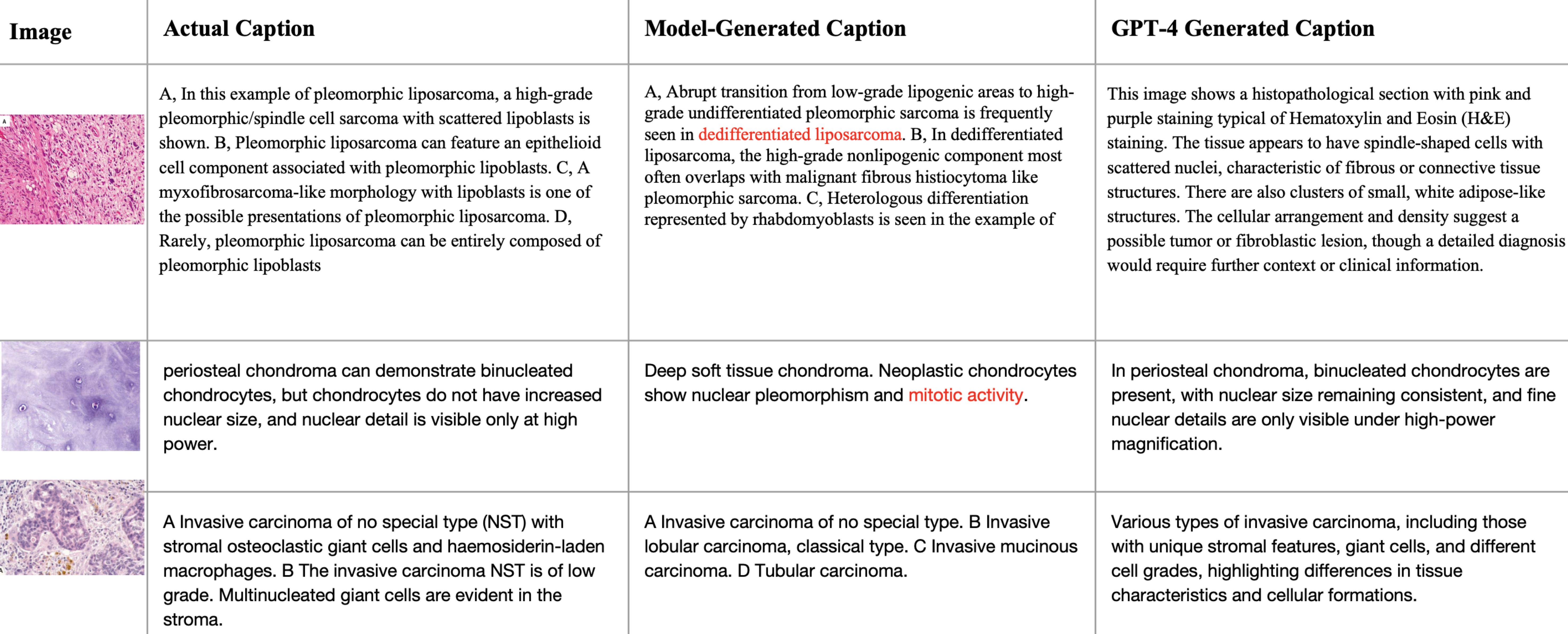}}
    \caption{Comparison of actual, model-generated, and GPT-4o generated captions for histopathology images. The left column shows the actual captions, followed by the captions generated by the proposed model in the middle column, and those generated by GPT-4o in the right column. The image demonstrates how each method describes histopathology features, including the accuracy of medical terminology and relevance to the image content. \cite{{8346302}}}
    \label{fig:result}
\end{figure}

In addition to the automatic evaluation metrics, we compared the performance of our model with GPT-4o, a state-of-the-art large language model, in terms of the relevance and quality of the generated captions as shown in Figure~\ref{fig:result}.  The comparison was based on several criteria, including:

\begin{itemize}
    \item Relevance to Image Content: How well the generated caption describes the features present in the histopathology image.
    \item Medical Accuracy: The correctness and medical appropriateness of the terminology used in the captions.
    \item Completeness: Whether the generated caption provides a comprehensive description of the key features in the image.
\end{itemize}

Our model outperformed GPT-4 in terms of medical accuracy and relevance to image content. While GPT-4o demonstrated strong general language generation abilities, it occasionally struggled with the domain-specific vocabulary and the contextual relevance required for histopathology images. On the other hand, our model, fine-tuned on the ARCH dataset, was specifically trained to handle medical terminology and pathology-related content, resulting in more accurate and contextually appropriate captions.

However, GPT-4o exhibited superior fluency and diversity in language generation, especially in scenarios where the image content was less complex. Our model, although more accurate in generating medical descriptions, sometimes produced captions that were more rigid or repetitive in structure. Despite these differences, both models demonstrated strong potential for generating captions for histopathology images, with our model showing a clear advantage in medical relevance and accuracy.

\subsection{Image-to-Text Alignment}

\begin{figure}[!ht]
    \centerline{\includegraphics[width=0.3\textwidth]{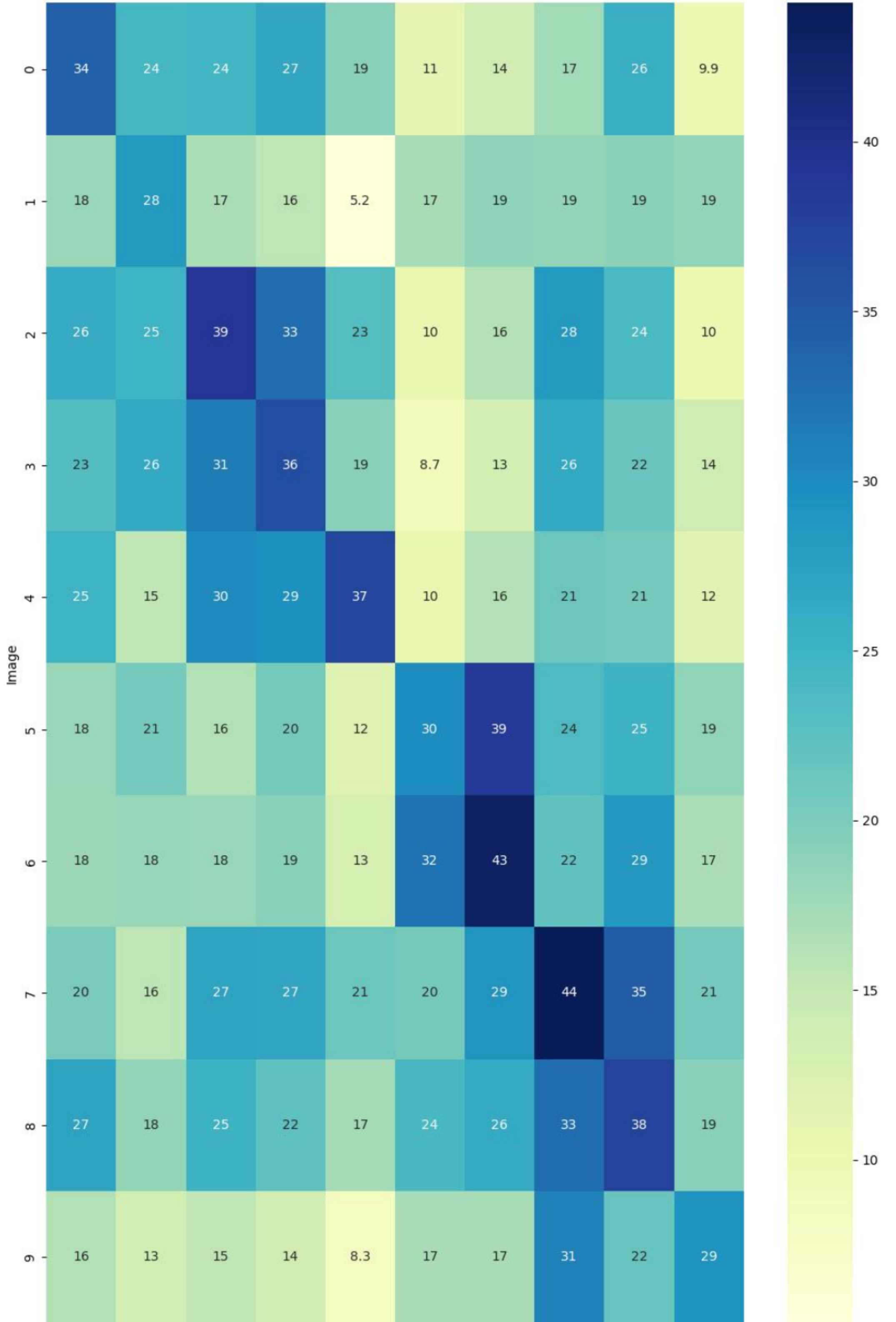}}
    \caption{Heatmap representation of the data matrix. Each cell value corresponds to the intensity of a particular variable across images, with the color gradient indicating the magnitude of the values. The color bar on the right shows the scale from low (green) to high (blue) values, ranging from 9.9 to 44. }
    \label{fig:heatmap}
\end{figure}

To further assess the quality of the generated captions, we visualized the alignment between images and their generated captions using a heatmap of image-to-text similarity as shown in Figure~\ref{fig:heatmap}. This heatmap plots the similarity between each generated caption and its corresponding image across different regions. The heatmap indicated that the model was generally effective in generating captions that aligned with the most prominent features in the images, such as tissue boundaries and obvious abnormalities.

However, the model struggled to generate accurate captions in cases where the images contained more complex or subtle features, such as fine tissue details or very small abnormalities. These cases resulted in less accurate or incomplete captions, as seen in the off-diagonal regions of the heatmap, which correspond to images with lower alignment to the generated captions.

\subsection{Ablation study}

An ablation study was conducted to assess the individual contributions of different components and design choices in our model. The aim was to better understand how each part of the model and training process impacts the overall performance. The ablation study was designed to isolate key features, such as the use of Vision Transformer (ViT), GPT-2, pretraining and fine-tuning on the specialized ARCH dataset. 

The following experiments were performed in the ablation study:

\subsubsection{ViT vs. CNN for Image Feature Extraction} 
One of the critical components of our model is the use of Vision Transformers (ViT) for image feature extraction. To assess the contribution of ViT, we compared it against a traditional Convolutional Neural Network (CNN) such as ResNet-50. The CNN model was fine-tuned on the ARCH dataset, while the ViT model followed the same preprocessing and fine-tuning procedures. Both models were then coupled with GPT-2 for caption generation. The ViT model outperformed the CNN model across all metrics. Specifically, the ViT model achieved higher ROUGE-1 and ROUGE-2 scores, as well as improved BLEU scores, indicating that ViT's attention mechanisms were more effective at capturing global image context and fine-grained details compared to the CNN's local receptive fields. On the other hand, the CNN model struggled with more complex visual relationships and long-range dependencies in histopathology images, which contributed to its lower performance in generating accurate and complete captions.

\subsubsection{Fine-Tuning with ARCH Dataset vs. Fine-Tuning with General Dataset}

To assess the benefit of using a domain-specific dataset like ARCH, we compared the performance of our model after fine-tuning on the ARCH dataset versus fine-tuning on a more general image-captioning dataset, such as MSCOCO or Flickr30k. The pre-trained ViT and GPT-2 models were fine-tuned on these two datasets separately, and the performance was evaluated using the same metrics (ROUGE, BLEU, human evaluation). Fine-tuning on the ARCH dataset resulted in substantially better performance across all evaluation metrics. The ROUGE-1 and ROUGE-2 scores were significantly higher for the ARCH fine-tuned model, indicating that the model better understood the domain-specific language and content of histopathology images. In contrast, the general dataset fine-tuned model struggled to generate accurate medical captions and failed to capture the nuanced terminology necessary for pathology-related tasks. The lack of medical domain knowledge caused the captions to be too general and sometimes irrelevant to the specific medical features in the images.

\section{Discussion}
In this study, we introduced a novel approach for captioning histopathology images, leveraging Vision Transformers (ViT) for image feature extraction and GPT-2 for text generation. Our approach, fine-tuned on the specialized ARCH dataset, demonstrates significant improvements over traditional models that rely on pre-trained networks like ImageNet. The ViT model's ability to capture long-range dependencies and intricate visual features is particularly valuable in pathology, where fine details such as tissue morphology and staining variations are critical for accurate diagnosis. Unlike traditional CNNs, which struggle with long-range dependencies, ViT excels in recognizing complex visual patterns, making it a better fit for analyzing pathology images.

Fine-tuning the pre-trained ViT and GPT-2 models on the ARCH dataset, which contains dense, domain-specific captions, allowed the model to understand medical terminology and pathology-related concepts. This domain adaptation significantly enhanced the relevance and accuracy of the generated captions compared to general-purpose datasets like MSCOCO. Additionally, the use of transfer learning from large-scale image-text datasets such as MSCOCO accelerated training and improved model performance, enabling the model to adapt to the specific requirements of histopathology.

Our approach shows strong potential for enhancing diagnostic capabilities in computational pathology, as the generated captions provide more accurate and contextually relevant descriptions of histopathological images. This makes it possible to improve the classification, segmentation, and retrieval of pathology images based on their content, leading to better diagnostic support for pathologists. By combining the power of vision and language models, the proposed model paves the way for more effective integration of visual and textual data, which is crucial for advancing the field of computational pathology.

\subsection{Limitations and Future Work}
Although the proposed model demonstrates significant improvements over traditional methods, there are several limitations that should be addressed in future work. One primary limitation is the model's reliance on the ARCH dataset, which is specialized for histopathology images. While this makes the model highly effective for pathology tasks, it limits its ability to generalize across other medical imaging fields, such as radiology or dermatology. Future research could involve integrating diverse datasets from different medical domains to increase the model’s generalizability. Expanding the dataset to include a wider variety of pathological conditions, tissue types, and staining techniques would further improve the robustness of the model and make it more applicable to real-world clinical settings.

Another limitation is the high computational cost associated with training transformer-based models, especially when working with multi-gigapixel whole-slide images. The model requires substantial computational resources for both training and inference, which may not be available in resource-constrained environments such as smaller hospitals or research labs. To address this, future work could focus on optimizing the model’s architecture to reduce its computational footprint \cite{CHI}. Techniques such as model pruning, knowledge distillation, or quantization could be explored to improve efficiency without sacrificing performance. Additionally, optimizing the inference pipeline to reduce processing time would be crucial for deploying the model in time-sensitive clinical scenarios.

To further enhance the model's scalability and versatility, future research could focus on integrating semi-supervised and unsupervised learning methods. Given the scarcity of annotated medical data, especially for rare diseases, semi-supervised techniques such as self-training or consistency regularization could help the model learn from unlabeled data, reducing the reliance on large annotated datasets. Unsupervised pre-training on large amounts of unlabeled histopathology data could also allow the model to capture rich, domain-specific feature representations, enabling it to generalize better to new, unseen datasets.

\section{Conclusion}

This study demonstrates the potential of deep learning approaches, specifically the combination of Vision Transformers (ViT) and GPT-2, for improving captioning tasks in computational pathology. By fine-tuning on the specialized ARCH dataset, our model is able to generate accurate, contextually relevant, and medically-specific captions for histopathology images, outperforming traditional models that rely on general-purpose datasets like ImageNet. Despite the promising results, the model’s dependence on domain-specific data, high computational requirements, and sensitivity to staining variations present notable challenges. Future work will focus on addressing these limitations by exploring techniques such as semi-supervised learning, data augmentation to handle staining variability, and model optimization for scalability in clinical settings. Our approach contributes to the broader field of computational pathology by enabling more effective use of multi-modal data \cite{You2024VRARWorkshop} (images and text) for tasks like tissue classification, object detection \cite{tomar2024knowledgeinfusedautomateddiseasediagnosis}, and medical image retrieval. The integration of domain-specific data with advanced deep learning techniques has the potential to significantly enhance diagnostic capabilities and improve healthcare outcomes, providing a valuable tool for pathologists and clinicians.

\bibliographystyle{unsrt} 

\bibliography{references}

\end{document}